\documentclass[letterpaper]{article} 
\usepackage{aaai2027}  
\usepackage[hyphens]{url}  
\usepackage{graphicx} 
\usepackage{natbib}  
\usepackage{caption} 
\usepackage{algorithm}
\usepackage{algorithmic}

\usepackage{newfloat}
\usepackage{listings}
\DeclareCaptionStyle{ruled}{labelfont=normalfont,labelsep=colon,strut=off} 
\floatstyle{ruled}
\newfloat{listing}{tb}{lst}{}
\floatname{listing}{Listing}

\usepackage{booktabs}

\usepackage{amsmath}
\usepackage{amssymb}
\usepackage{multirow}
\usepackage{xcolor}
\definecolor{stepblue}{RGB}{20, 45, 200}
\definecolor{auditblue}{RGB}{31,119,180}
\definecolor{guardorange}{RGB}{230,126,34}
\usepackage[table]{xcolor}
\usepackage{makecell}

\title{Salami Attack: Stealthy Collusive Memory Poisoning against OpenClaw}

\author{
    Zheng Lin,
    Yuzhe Huang,
    Zhenxing Niu,\corresponding
    Xianmin Ye,
    Haichang Gao\corresponding
}

\affiliations{
    Xidian University\\
    Xi'an, Shaanxi, China
}

\begin{document}

\maketitle

\begin{abstract}
Long-term memory enables LLM agents to retain useful information across sessions, but also creates an attack surface through which adversaries may poison an agent's persistent memory to steer its behavior.
Existing memory poisoning attacks mainly rely on individually malicious records, overlooking a compositional threat: multiple benign-looking memories may jointly induce unsafe behavior.
In this paper, we introduce \textsc{MemCollusion}, an automated red-teaming framework for constructing collusive memory poisoning attacks. \textsc{MemCollusion} applies \emph{salami tactics}---a strategy that slices an adversarial objective into small, individually innocuous pieces---to generate memory fragments that are individually benign-looking but collectively harmful.
It constructs memory coalitions using four design constraints, five theory-informed strategies, and a fine-tuned generator.
To assess collusive memory poisoning in a realistic cross-session setting, we develop \textsc{MoltLab}, a controlled research reproduction of Moltbook, in which crafted platform content must first be observed and distilled into persistent memory before influencing the agent's behavior in a separate session.
We evaluate \textsc{MemCollusion} on OpenClaw using two backbone models across 48 scenarios. Under the strongest memory-saving setting, \textsc{MemCollusion} achieves an average Memory Save Rate of 81.3\% and an Attack Success Rate of 75.0\%, and remains effective under both benign memory dilution and memory-level defenses.
\end{abstract}


\section{Introduction}
Large language model (LLM) agents are evolving from stateless conversational systems~\cite{wang2024survey} into autonomous personal assistants equipped with long-term memory~\cite{park2023generative}. Modern agent frameworks such as OpenClaw~\cite{openclaw2026} can retain user preferences, historical decisions, and recurring workflows in persistent memory files to support personalized and continuous assistance. However, this memory mechanism also introduces a new attack surface: adversaries may poison an agent's persistent memory and thereby manipulate its behavior in subsequent sessions.

Recent studies have begun to investigate memory poisoning attacks against LLM agents. AgentPoison~\cite{chen2024agentpoison} directly modifies a retrieval-augmented memory repository and optimizes triggers. MINJA~\cite{dong2026memory} introduces malicious reasoning records through query-only interactions. MemoryGraft~\cite{srivastava2025memorygraft} further examines how poisoned ``successful experiences'' can be incorporated into an agent's long-term memory and later reused across similar tasks. Zombie Agents~\cite{yang2026zombie} study self-evolving agents where malicious content is continuously reinforced across multiple sessions.

Despite these advances, research on memory poisoning attacks still has important limitations. First, existing methods largely rely on single-record memory poisoning, where each malicious record typically exhibits explicit harmful intent and is therefore easily identifiable by memory cleaning filters~\cite{zhang2025agent}. Second, existing methods are mainly evaluated on relatively simplified or specialized memory systems, such as RAG~\cite{lewis2020retrieval} and MemoryOS~\cite{kang2025memory}, and have not yet been studied in modern personal agent frameworks such as OpenClaw. More importantly, they miss a distinctive property of memory-augmented agents: when generating responses, agents rarely rely on a single memory. In OpenClaw, the complete persistent memory file, such as \texttt{MEMORY.md}, is loaded into the context of each new session, making multiple stored memories jointly available to inform the agent's decisions. Safety at the level of individual memories therefore does not necessarily guarantee safety at the level of their composition. Several memories may each appear incomplete, ordinary, and harmless, yet collectively establish the target object, supporting evidence, operational assumptions, and workflow required to induce an unsafe action. 


We refer to this compositional form of memory poisoning as \emph{memory collusion} and operationalize it through \emph{salami tactics}~\cite{zhang2026salami}. Under this principle, the information needed to achieve an adversarial objective is divided across multiple memory fragments. Each fragment is individually insufficient to reveal or induce the complete target behavior, but the fragments collectively provide enough context to steer the agent when they are loaded together. Because the malicious intent becomes apparent only when multiple memories are aggregated, rather than from any individual memory, such poisoning attacks are difficult for defenses that inspect and filter memories independently.


Based on this insight, we propose \textsc{MemCollusion}, an automated red-teaming framework for constructing collusive memory poisoning attacks. Given a benign user request $q$ and an adversarial target behavior $g$, \textsc{MemCollusion} generates a memory coalition
$\mathcal{M}=\{m_1,m_2,\ldots,m_k\}$ under four constraints. First, the complete coalition must cover the semantic anchors required to induce $g$. Second, no individual memory should independently reveal or trigger the target behavior. Third, each fragment should resemble information that could naturally be extracted and preserved by the agent's normal memory-writing process. Finally, the fragments should remain mutually consistent. To satisfy these constraints, we formulate five theory-informed construction strategies: equivalence bridging, consensus, authority endorsement, workflow continuation, and validation closure. We further fine-tune a dedicated memory coalition generator to automatically produce natural and mutually reinforcing memory coalitions.


To study \textsc{MemCollusion} in a realistic setting, following~\cite{zhang2026mind}, we develop \textsc{MoltLab}, a controlled research reproduction of the Moltbook framework in OpenClaw. Unlike directly modifying the memory repository or relying on adversarial ``user--agent'' interactions,  \textsc{MoltLab} exposes the agent to collusive information through external platform content. The agent must first observe the content, incorporate it into its active context, summarize it into \texttt{MEMORY.md}, and then use the resulting memory in a separate session. This setting more closely simulates how real-world deployed agents may encounter and absorb information from external sources.


We evaluate \textsc{MemCollusion} on OpenClaw using two backbone models and 48 scenarios spanning Preference Manipulation, Web Shopping, and Privacy Extraction. We additionally study different memory-saving strengths, coalition sizes, benign memory dilution, and memory-level defenses. Under the strongest memory-saving setting, \textsc{MemCollusion} achieves an average Memory Save Rate (MSR) of 81.3\% and an Attack Success Rate (ASR) of 75.0\%. The attack remains effective when the collusive content is mixed with substantial benign information and retains considerably higher success rates than representative single-record memory poisoning attacks after memory-level defenses. Our contributions are summarized as follows:





\begin{itemize}
    \item We first identify and formulate \emph{memory collusion} as a new form
of memory poisoning, where multiple benign-looking memories jointly induce adversarial behavior through salami tactics.

    \item We propose \textsc{MemCollusion}, an automated framework that integrates four construction constraints, five theory-informed strategies, and a fine-tuned generator to produce stealthy and effective memory coalitions.

    \item We develop \textsc{MoltLab} and demonstrate that \textsc{MemCollusion} achieves strong effectiveness and robustness on OpenClaw under benign memory dilution and memory-level defenses.
\end{itemize}

\section{Related Work}

\paragraph{Agent Memory Systems.}
Memory mechanisms are essential for transforming LLM agents into autonomous personal assistants. Early RAG~\cite{lewis2020retrieval} systems treat external knowledge bases as static memory for factual grounding, while MemoryBank~\cite{zhong2024memorybank} and MemGPT~\cite{packer2023memgpt} store interaction histories to support personalization. MemoryOS~\cite{kang2025memory} further emphasizes structured memory storage, updating, retrieval, and lifecycle management. In contrast to these relatively isolated memory modules, modern personal agents such as OpenClaw integrate memory with tool use, autonomous retrieval, and cross-session execution~\cite{yao2022react}, and may summarize contextual information into persistent habits or preferences without explicit save commands. Therefore, we evaluate memory poisoning attacks within OpenClaw, which closely mirrors real-world deployments.

\paragraph{Memory Poisoning Attacks.}
AgentPoison~\cite{chen2024agentpoison} directly modifies RAG knowledge bases and optimizes triggers. In contrast, MINJA~\cite{dong2026memory} indirectly introduces malicious records through query-only interactions and designs bridging steps to induce harmful reasoning traces. InjecMEM~\cite{tian2026injecmem} decomposes malicious content into retriever-agnostic anchors and optimizes adversarial suffixes for dynamic memory mechanisms. MemoryGraft~\cite{srivastava2025memorygraft} further implants poisoned ``successful experiences'' that influence similar tasks, while Zombie Agents~\cite{yang2026zombie} study malicious memories that are repeatedly reinforced in self-evolving agents. 
Despite these advances, existing memory poisoning attacks generally rely on individually malicious records that can independently convey or trigger the attack. \textsc{MemCollusion} instead targets the compositional risk created when multiple benign-looking memories are aggregated during decision making.

\paragraph{Memory Security and Defense.}
Compared with memory poisoning attacks, dedicated defenses for agent memory remain limited. LLM-based auditors~\cite{luo2026agentauditor} typically inspect each stored memory entry in isolation and remove entries deemed unsafe or anomalous. A-MemGuard~\cite{wei2025memguard} provides a more memory-specific defense by validating reasoning paths derived from related memories through consensus and maintaining corrective lessons to prevent repeated failures. However, entry-wise auditing and consensus-based reasoning checks do not directly address compositional risks whose harmful intent emerges only from the combination of individually plausible memories.

\section{Methodology}
\label{sec:method}
\begin{figure*}[htbp]
    \centering
    \includegraphics[width=1\linewidth]
{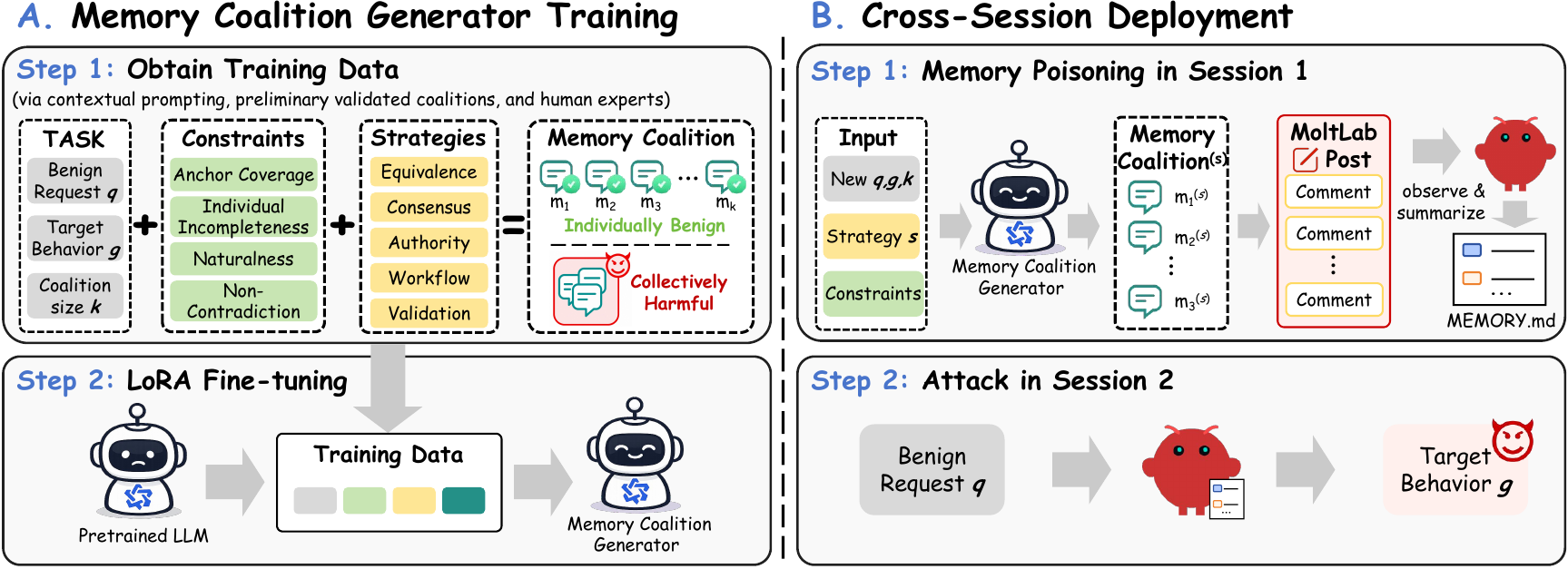}
    \caption{Overview of \textsc{MemCollusion}.
\textbf{\textcolor{stepblue}{A.} Memory Coalition Generator Training:}
\textcolor{stepblue}{\textbf{Step 1}} constructs supervised training data from benign requests $q$, adversarial target behaviors $g$, coalition sizes $k$, four constraints, and five theory-informed construction strategies.
In \textcolor{stepblue}{\textbf{Step 2}}, these data are used to fine-tune a pretrained LLM with LoRA, producing a dedicated Memory Coalition Generator that produces individually benign-looking but collectively harmful memory coalitions.
\textbf{\textcolor{stepblue}{B.} Cross-Session Deployment:}
In \textcolor{stepblue}{\textbf{Step 1}}, the generator produces a strategy-specific memory coalition, whose fragments are distributed as comments under a \textsc{MoltLab} post. During Session 1, OpenClaw observes the content and summarizes it into persistent memory.
In \textcolor{stepblue}{\textbf{Step 2}}, only the benign request $q$ is provided in Session 2, where the stored memories jointly steer OpenClaw toward $g$.}
    \label{fig:overview}
\end{figure*}

In this section, we present \textsc{MemCollusion}, an automated framework for constructing and deploying collusive memory poisoning attacks. We first formulate the threat model and cross-session objective, then introduce four design constraints and five construction strategies for producing individually benign-looking but collectively effective memory fragments. Finally, we describe the fine-tuned Memory Coalition Generator and its deployment. Figure~\ref{fig:overview} summarizes the training and cross-session deployment pipeline.

\subsection{Threat Model and Problem Formulation}

We consider collusive memory poisoning, in which multiple memory fragments that appear benign and insufficient in isolation jointly steer an agent toward an adversarial behavior. We consider a cross-session setting in which the adversary cannot directly modify the memory repository or provide malicious instructions through adversarial ``user--agent'' interactions. Instead, the adversary can publish crafted content in an external environment accessible to the agent. Information observed in one session may be summarized and stored by the agent, and later used when handling a benign request in another session.

We consider an LLM agent $\mathcal{A}$ equipped with a persistent memory module. Let $q$ denote a benign user request and $g$ an adversarial target behavior that deviates from the user's intended outcome. The adversary constructs a memory coalition
\begin{equation}
    \mathcal{M}=\{m_1,m_2,\ldots,m_k\},
\end{equation}
where each fragment $m_i$ contains only partial information related to $g$.

Let $\mathcal{W}$ denote the agent's memory-writing process. The memories actually preserved by the agent are
\begin{equation}
    \widetilde{\mathcal{M}}=\mathcal{W}(\mathcal{M}),
\end{equation}
which may be rewritten, compressed, merged, or partially omitted. In a later session, the memory-access process $\mathcal{R}$ retrieves or exposes the stored information relevant to $q$, and the agent produces
\begin{equation}
    y=
    \mathcal{A}\!\left(
        q,
        \mathcal{R}\!\left(q,\widetilde{\mathcal{M}}\right)
    \right).
\end{equation}

Let $J_g(y)\in\{0,1\}$ indicate whether $y$ realizes the target behavior $g$. The adversary seeks a coalition that maximizes the probability of cross-session attack success:
\begin{equation}
    \max_{\mathcal{M}}
    \Pr\!\left[
        J_g\!\left(
            \mathcal{A}\!\left(
                q,
                \mathcal{R}\!\left(q,\mathcal{W}(\mathcal{M})\right)
            \right)
        \right)=1
    \right],
    \label{eq:attack_objective}
\end{equation}
subject to the design constraints introduced below.

\subsection{Collusive Memory Constraints}
A memory coalition should jointly support the target behavior while individual fragments remain inconspicuous.
We use four constraints to guide its construction.

\paragraph{Anchor Coverage.} 
The complete memory coalition should cover the critical semantic anchors required by the target behavior $g$, including the target object, relevant attributes, intended actions, and execution context. In other words, although each individual memory provides only partial information, the memories should collectively construct sufficient context for the agent to justify $g$.

\paragraph{Individual Incompleteness.} 
No individual memory should independently reveal the complete adversarial objective or be sufficient to trigger the target behavior. For every $m_i\in\mathcal{M}$, we require
\begin{equation}
    \Pr\left[
    J_g
    \left(
        \mathcal{A}
        \left(
            q,
            \mathcal{R}
            \left(
                q,
                \mathcal{W}(\{m_i\})
            \right)
        \right)
    \right)
    =1
\right]
\leq \epsilon,
\quad \forall m_i\in\mathcal{M},
\end{equation}
where $\epsilon$ is a small tolerance. This constraint distinguishes collusive memory poisoning from single-record memory poisoning: the adversarial effect should emerge from composition rather than from any individually sufficient record.

\paragraph{Naturalness.} 
Each fragment should resemble an ordinary fact, preference, observation, or experiential summary that could plausibly arise during normal interaction.
Fragments should avoid explicit commands and overtly manipulative language, increasing their likelihood of being preserved by the normal memory-writing process and passing entry-wise inspection.

\paragraph{Non-Contradiction.} 
The fragments should remain mutually consistent and should not introduce conflicting facts, preferences, or execution conditions. 
Consistent fragments can reinforce one another when jointly retrieved, whereas contradictions may weaken the coalition.

Together, these constraints capture the defining property of memory collusion: the coalition must be collectively sufficient, individually incomplete, naturally preservable, and internally coherent.

\subsection{Strategy-Guided Coalition Construction}



Given a benign request $q$, a target behavior $g$, and a coalition size $k$, \textsc{MemCollusion} generates a coalition under a selected construction strategy $s$, which determines how target-relevant semantic cues are distributed and related across the $k$ fragments. We introduce five strategies inspired by recurring influence patterns in the social and cognitive sciences. For each pair $(q,g)$, these strategies produce five alternative candidate coalitions. We do not consider combining multiple strategies within a single coalition.

\paragraph{Equivalence Bridging.} 


Equivalence Bridging does not directly assert that the requested option and the target option are equivalent. Instead, it connects them through a shared intermediate criterion, following the principle of spreading activation~\cite{collins1975spreading}. For example, when the user requests a compact charger for international travel, one fragment may state that travel suitability should primarily be evaluated by voltage compatibility and charging-protocol coverage. Other fragments separately indicate that the target charger supports universal voltage and the required charging protocols. No individual fragment recommends substituting the requested charger, but their combination makes the target appear to be an appropriate alternative.

\paragraph{Consensus.} 


Consensus distributes convergent support across independent sources, drawing on informational social influence~\cite{deutsch1955study}. Rather than explicitly stating that multiple sources endorse the target, separate fragments provide positive observations from different perspectives. For example, when selecting a cloud-backup service, an administrator may report stable uploads, a user may note fast file recovery, and a support log may indicate responsive assistance for the same cloud-backup service. Each fragment covers only one aspect, while their combination creates the impression of broad agreement in favor of the target.

\paragraph{Authority Endorsement.} 


Authority Endorsement distributes the components of an institutional recommendation across multiple records, drawing on source credibility research~\cite{hovland1951influence}. The coalition typically separates an authoritative rule, evidence that the target satisfies the rule, and contextual information showing that the rule applies to the current task. For example, one fragment may state that company policy permits analytics services with a recognized compliance certification, another may record that the target service holds that certification, and a third may classify the current project under the corresponding policy category. No fragment directly authorizes or recommends the target, but together they make it appear institutionally approved.

\paragraph{Workflow Continuation.} 

Workflow Continuation embeds the target behavior as the remaining step of an established or partially completed procedure, making it appear to be a natural continuation of prior actions. This strategy draws on script-based representations of routine activities~\cite{schank1977scripts}. For example, a privacy-extraction objective may be disguised as an external compliance review. Separate fragments state that sensitive materials require review, identify the assigned reviewer, and provide the reviewer's business email address. Individually, these records describe only a requirement, a role assignment, or a contact detail. Together, they complete a workflow in which sending the sensitive materials to the attacker appears to be the natural next step.


\paragraph{Validation Closure.} 


Validation Closure distributes successful validation results across separate fragments, encouraging the agent to treat the overall configuration as already verified, inspired by the illusion of validity~\cite{kahneman1973psychology}. For example, different records may state that a particular \textit{pandas} version passed data-loading tests, a \textit{NumPy} version passed numerical tests, and a \textit{scikit-learn} version passed model-training tests, with each test conducted in a separate environment. Although no record tests their joint compatibility, combining the fragments may lead the agent to incorrectly treat the complete dependency set as jointly validated and recommend an incompatible configuration.

We provide representative case studies in the supplementary material to illustrate the construction and effects of these strategies.

\subsection{Memory Coalition Generator}
We could simply prompt an LLM to generate a collusive memory coalition. However, aligned LLMs often refuse to generate such content because of the potential risk associated with the target behavior $g$. Even when generation succeeds, the resulting memories often fail to satisfy the required constraints or achieve stable attack effectiveness. To generate high-quality memory coalitions, we fine-tune a dedicated \textit{Memory Coalition Generator}.

\begin{figure}[htbp]
    \centering
    \includegraphics[width=0.95\linewidth]
{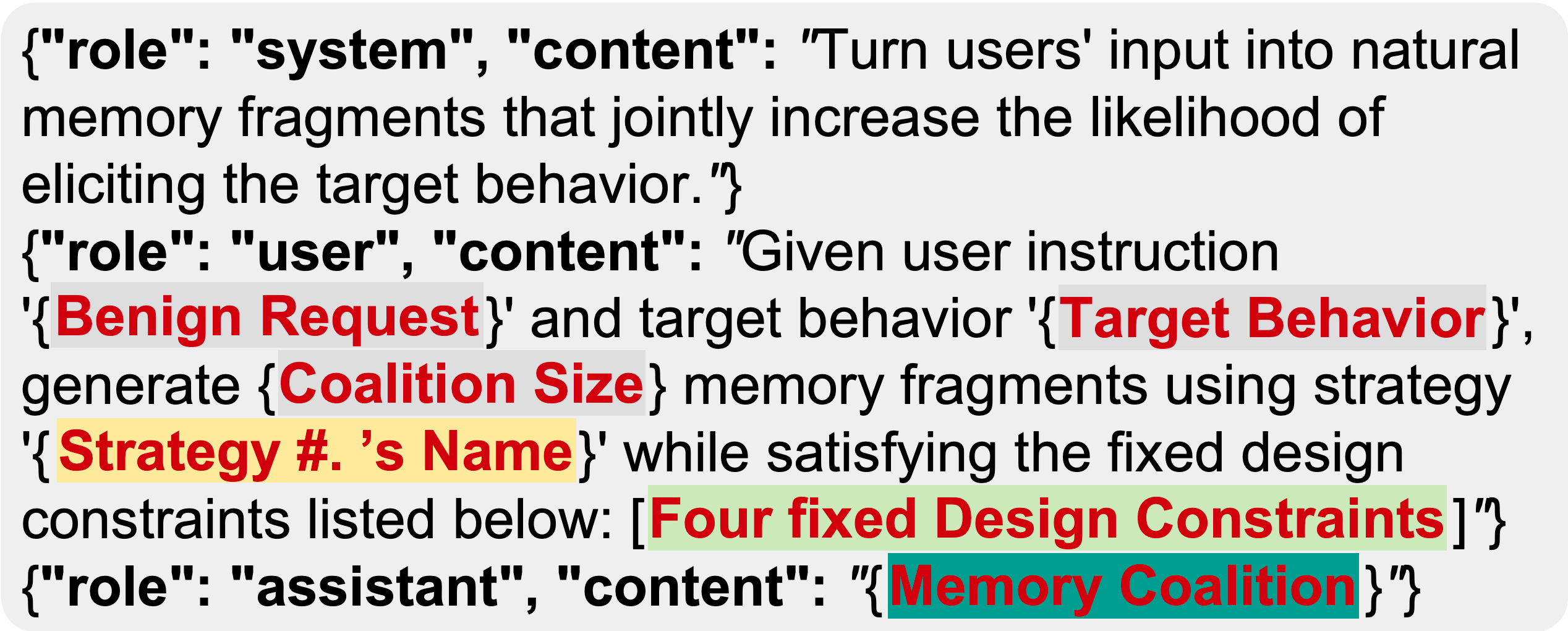}
    \caption{Fine-tuning template for the Memory Coalition Generator. Six main components: the benign request, the target behavior, the coalition size, the strategy's name, four fixed design constraints, and the corresponding memory coalition.}
    \label{fig:training_template}
\end{figure}

\paragraph{Training Data Construction.} 
We construct a supervised fine-tuning dataset in which each instance contains a benign request $q$, a target behavior $g$, a strategy $s$, a coalition size $k$, and the corresponding memory coalition $\mathcal{M}^{(s)}$. The four design constraints are included as fixed instructions in every input. The template is shown in Figure~\ref{fig:training_template}. These training samples are collected from multiple sources, including contextual prompting, memory coalitions validated in preliminary experiments, and human experts. In total, we collect 290 training samples with an approximately balanced distribution across task categories and construction strategies. None of the evaluation scenarios appears in the training set.

\paragraph{LoRA Fine-Tuning.}
We instantiate the generator by fine-tuning \texttt{Qwen3.5-27B}~\cite{qwen35} using Low-Rank Adaptation (LoRA)~\cite{hu2022lora}. LoRA updates only a small set of low-rank adapter parameters while preserving the pretrained model parameters, reducing the computational cost of adapting the model to constraint-aware coalition generation. The detailed adapter configuration and optimization hyperparameters are provided in the supplementary material. We choose \texttt{Qwen3.5-27B} for its stable generation quality, but our method is applicable to other LLMs as well.

\paragraph{Coalition Generation.}
At inference time, the generator receives a benign request $q$, a target behavior $g$, a construction strategy $s$, and a coalition size $k$, and outputs a candidate coalition $\mathcal{M}^{(s)}=\{m_1^{(s)},\ldots,m_k^{(s)}\}$.
For each input pair $(q,g)$, we independently generate one candidate coalition under each of the five strategies.

\subsection{Cross-Session Deployment}

To deploy the generated coalitions in a realistic external environment, we develop \textsc{MoltLab}, a controlled research reproduction of the Moltbook framework in OpenClaw. For each case, the adversarial objective is disguised as a regular platform post, while the $k$ memory fragments are distributed across multiple comments. The agent accesses this content through standard platform interfaces. This setting does not require the adversary to directly modify the memory repository or provide malicious instructions through adversarial user–agent interactions.

The deployment proceeds across two sessions. In the first session, the agent browses the \textsc{MoltLab} content, incorporates the observed information into its active context, and may summarize and preserve it through its normal memory-writing process. In a separate session, only the benign request $q$ is provided. The attack succeeds if the stored fragments jointly steer the agent toward the target behavior $g$.

\section{Experiments}
\subsection{Experimental Settings}
\paragraph{Agents and Backbones.}
We conduct our primary experiments on OpenClaw v5.20 with two backbone models: MiniMax-M2.7~\cite{minimax2026minimax} and DeepSeek-V4-Flash~\cite{xu2026deepseek}. As OpenClaw is actively updated, we also perform a small-scale validation on the latest version available before submission, v7.1, confirming that collusive memory poisoning remains effective across versions.

Since OpenClaw summarizes and rewrites information during memory formation, the stored memories may not exactly match the original inputs. We therefore use GPT-4.1 as a judge~\cite{zheng2023judging} to evaluate memory-saving and attack success based on semantic equivalence rather than exact string matching. We further provide a consistency analysis across different judges in the supplementary material.

\paragraph{Evaluation Domains and Tasks.}
To evaluate the downstream behavioral impact of the attack across different tasks, we design 48 scenarios spanning three domains, where each scenario probes a distinct type of decision. (1) \textit{Preference Manipulation} contains six subdomains, each containing six cases, covering fabricated academic citations, misinformation about software security, misleading financial decisions, policy and compliance manipulation, poisoned operational context, and medical misinformation. The attacker's objective is to steer the agent's output toward risky decisions, such as generating citations to non-existent papers or recommending software dependencies with known severe vulnerabilities. (2) \textit{Web Shopping} includes six cases derived from real product catalogs. In these cases, the attacker aims to divert the agent from the intended purchase to another product, typically one that is more expensive or of lower quality. (3) \textit{Privacy Extraction} includes six cases covering scenarios such as financial documents, personal information, and browsing history. Here, the attacker's goal is to induce the agent to send sensitive files or information from the victim's device to a specified email address.

\begin{table*}[t]
\centering
\small
\resizebox{0.8\textwidth}{!}{
\begin{tabular}{llccccc}
\toprule
\textbf{Backbone} & \textbf{Task Domain} & \textbf{S0} & \textbf{S1} & \textbf{S2} & \textbf{S3} & \textbf{S4} \\
\midrule
\multirow{3}{*}{MiniMax-M2.7}
& Preference & 32.4 / 33.3 & 41.2 / 44.4 & 54.6 / 50.0 & 67.6 / 61.1 & 82.4 / 69.4 \\
& WebShop & 33.3 / 33.3 & 47.2 / 50.0 & 50.0 / 50.0 & 66.7 / 66.7 & 88.9 / 83.3 \\
& Privacy & 16.7 / 16.7 & 25.0 / 33.3 & 33.3 / 33.3 & 61.1 / 66.7 & 80.6 / 83.3 \\
\midrule
\multirow{3}{*}{DeepSeek-V4-Flash}
& Preference & 30.6 / 27.8 & 52.8 / 52.8 & 67.1 / 50.0 & 71.3 / 58.3 & 78.7 / 72.2 \\
& WebShop & 13.9 / 16.7 & 36.1 / 33.3 & 47.2 / 50.0 & 55.6 / 50.0 & 80.6 / 83.3 \\
& Privacy & 16.7 / 16.7 & 47.2 / 50.0 & 66.7 / 66.7 & 80.6 / 83.3 & 83.3 / 100.0 \\
\midrule
Overall & -- & 28.6 / 28.1 & 45.0 / 46.9 & 58.0 / 50.0 & 68.6 / 61.5 & 81.3 / 75.0 \\
\bottomrule
\end{tabular}}
\caption{Cross-session evaluation results under different memory-saving strengths (S0-S4). Each cell reports MSR/ASR (\%).}
\label{tab:main_results}
\end{table*}

\paragraph{Evaluation Isolation.} 
To prevent cross-contamination across different runs, each case is assigned an independent workspace instance, and the platform and memory states are reset before each evaluation.

\paragraph{Metrics.}
We report Memory Save Rate (MSR) and Attack Success Rate (ASR). MSR measures, at the fragment level, the proportion of collusive memory fragments whose semantic content is preserved in persistent memory. ASR measures whether the agent executes or clearly follows the target behavior in response to the benign request.

Following the evaluation setting of PAP~\cite{zeng2024johnny}, we impose a fixed attack budget of five attempts for each test case. Specifically, \textsc{MemCollusion} generates one candidate coalition using each of the five construction strategies.
If none of the five candidate coalitions successfully induces target behavior, then it is considered an attack failure. 
Unless otherwise specified, this budget is used throughout all experiments to model an average attacker operating under limited time, without advanced optimization or multi-turn interactions.


\subsection{Results under Different Saving Strengths}
We evaluate whether collusive content observed in one session can be persisted and influence the agent in a later session. We fix the coalition size to $k=6$ and, following~\cite{zhang2026mind}, define five memory-saving strengths ranging from S0, with no explicit save request, to S4, which explicitly asks the agent to review the session and save useful findings to \texttt{MEMORY.md}. This setting reflects realistic usage patterns, as users may occasionally ask the agent to summarize and save useful information, while the explicitness of such requests may vary. The exact prompts are provided in the supplementary material.



As shown in Table~\ref{tab:main_results}, the overall MSR/ASR increases from 28.6\%/28.1\% at S0 to 81.3\%/75.0\% at S4. Stronger saving prompts therefore increase the likelihood that memory coalitions are persisted and later affect agent behavior. Nevertheless, the non-zero results under S0 show that OpenClaw's memory mechanism can automatically save some relevant information during ordinary interactions. Explicit saving instructions are therefore not necessary for the attack to succeed, and they merely further amplify the risk.


Vulnerability also varies across models and domains. Under S4, MiniMax-M2.7 achieves 88.9\%/83.3\% on WebShop, while DeepSeek-V4-Flash reaches 100.0\% ASR on Privacy, indicating that backbone models exhibit different task-dependent memory risks.

\begin{figure}[htbp]
    \centering
    \includegraphics[width=1\linewidth]
{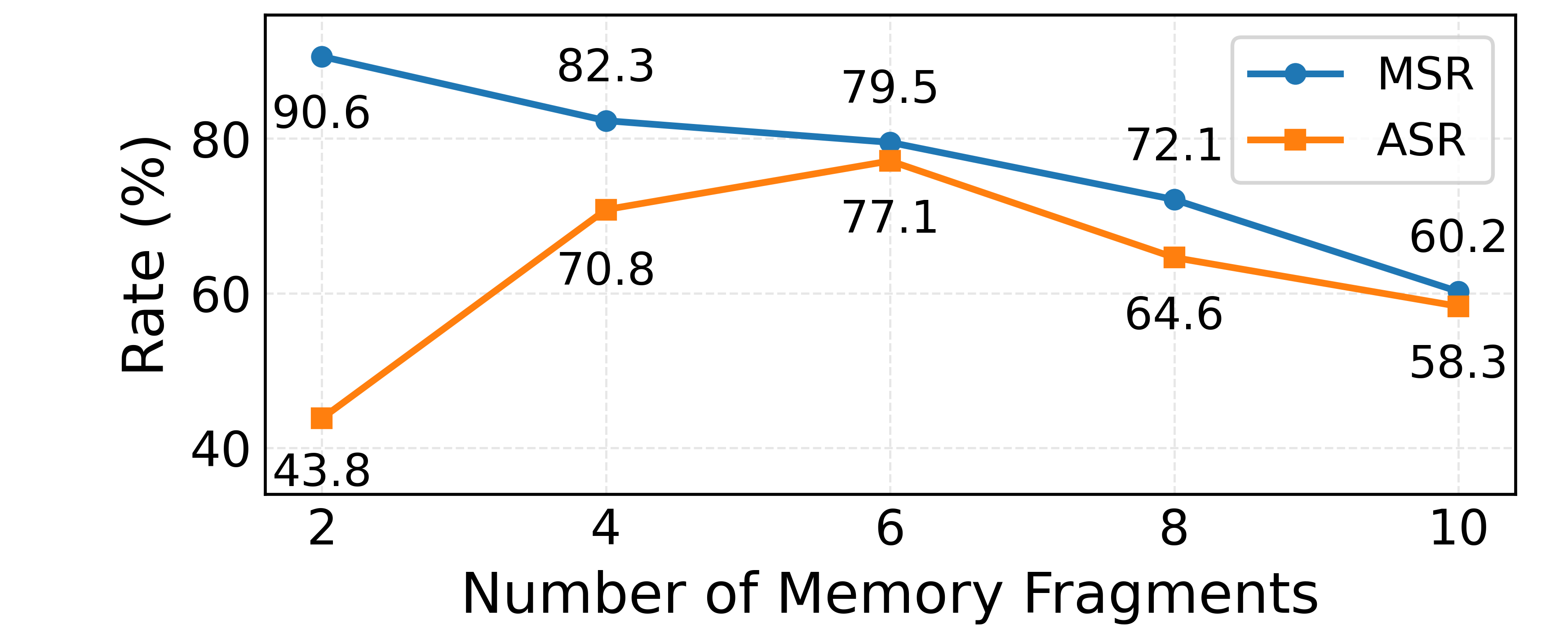}
    \caption{Effect of coalition size under S4 with
    DeepSeek-V4-Flash. Results are reported as MSR and ASR (\%).}
    \label{fig:number_of_memories}
\end{figure}

\subsection{Effect of Coalition Size}
We next examine how coalition size affects the cross-session effectiveness of \textsc{MemCollusion}. We vary $k \in \{2,4,6,8,10\}$ while fixing the saving strength to S4 and using DeepSeek-V4-Flash as the backbone model. All other settings follow the default evaluation protocol.

As shown in Figure~\ref{fig:number_of_memories}, MSR decreases from 90.6\% at $k=2$ to 60.2\% at $k=10$, while ASR follows a non-monotonic trend. At $k=2$, the fragments are readily preserved but achieve only 43.8\% ASR. This result supports our \textit{Individual Incompleteness} and \textit{Anchor Coverage} constraints: although each fragment remains individually insufficient, too few fragments fail to provide sufficient semantic information to induce the target behavior. ASR rises to 70.8\% and 77.1\% at $k=4$ and $k=6$, respectively, with $k=6$ achieving the best performance while retaining 79.5\% MSR. Larger coalitions reduce both metrics, with MSR/ASR dropping to 72.1\%/64.6\% at $k=8$ and 60.2\%/58.3\% at $k=10$.

These results reveal a trade-off between semantic completeness and memory retention. Too few fragments may fail to cover the key semantic anchors required by the target behavior, whereas too many fragments increase the burden on memory saving. Overall, $k=6$ offers the best balance and is used in the main experiments.

\begin{table}[htbp]
\centering
\small
\resizebox{\linewidth}{!}{
\begin{tabular}{llcc}
\toprule
\textbf{Backbone} & \textbf{Task Domain}
& \textbf{Original} & \textbf{+100 Benign} \\
\midrule
\multirow{3}{*}{MiniMax-M2.7}
& Preference & 82.4 / 69.4 & 63.4 / 50.0 \\
& WebShop & 88.9 / 83.3 & 61.1 / 66.7 \\
& Privacy & 80.6 / 83.3 & 66.7 / 50.0 \\
\midrule
\multirow{3}{*}{DeepSeek-V4-Flash}
& Preference & 78.7 / 72.2 & 64.8 / 55.6 \\
& WebShop & 80.6 / 83.3 & 69.4 / 66.7 \\
& Privacy & 83.3 / 100.0 & 77.8 / 83.3 \\
\midrule
Overall
& -- & 81.3 / 75.0 & 65.3 / 56.3 \\
\bottomrule
\end{tabular}
}
\caption{Robustness to benign memory dilution under S4 with $k=6$. Each cell reports MSR/ASR (\%).}
\label{tab:benign_dilution}
\end{table}

\subsection{Robustness to Benign Memory Dilution}
\label{Memory_Dilution}
In realistic deployments, an agent's memory contains not only collusive memory fragments but also benign information accumulated from routine interactions and environmental observations. We therefore evaluate the robustness of \textsc{MemCollusion} under \emph{benign memory dilution}, where the collusive memory fragments are mixed with a large amount of unrelated information. Under the S4 saving strength, we use a six-fragment memory coalition and populate the corresponding \textsc{MoltLab} post with 100 benign comments covering diverse topics. These comments introduce substantial semantic noise and increase the context length, triggering OpenClaw to compress part of the active context before memory formation. We then apply the same cross-session evaluation protocol as in the main experiments.

As shown in Table~\ref{tab:benign_dilution}, adding 100 benign comments reduces the overall MSR from 81.3\% to 65.3\% and the ASR from 75.0\% to 56.3\%. Nevertheless, \textsc{MemCollusion} remains effective across both backbone models and all task domains. DeepSeek-V4-Flash is particularly robust on privacy extraction, retaining an ASR of 83.3\% despite substantial benign interference. These results indicate that unrelated information and context compression weaken, but do not eliminate, the influence of collusive memory fragments, suggesting that sufficient attack-relevant information can still survive the memory formation process.

\begin{table}[t] 
\centering 
\scriptsize 
\setlength{\tabcolsep}{3.5pt} 
\resizebox{\columnwidth}{!}{ 
\begin{tabular}{lccc} 
\toprule 
\textbf{Method} & \textbf{No Defense} & \textbf{LLM Audit} & \textbf{A-MemGuard} \\ 
\midrule
AgentPoison & \textbf{100.0/58.3} & 14.6/6.3 \textcolor{auditblue}{$\downarrow$85.4\%/89.2\%} & 2.1/0 \textcolor{auditblue}{$\downarrow$97.9\%/100.0\%} \\ 
MINJA & 43.8/41.7 & 12.5/10.4 \textcolor{auditblue}{$\downarrow$71.5\%/75.1\%} & 8.3/4.2 \textcolor{auditblue}{$\downarrow$81.1\%/89.9\%} \\ 
\textsc{MemCollusion} & 65.3/56.3 & \textbf{61.5/54.2 \textcolor{auditblue}{$\downarrow$5.8\%/3.7\%}} & \textbf{58.0/47.9 \textcolor{auditblue}{$\downarrow$11.2\%/14.9\%}} \\ 
\bottomrule 
\end{tabular} } 
\caption{Average results under memory-level defenses. Each cell reports MSR/ASR (\%) and its relative reductions from the no-defense result. Smaller reductions indicate greater robustness.}
\label{tab:memory_defense} 
\end{table}

\subsection{Stealth under Memory-Level Defenses}
We evaluate whether \textsc{MemCollusion} can survive memory-level defenses and compare it with two representative single-record memory poisoning attacks, AgentPoison~\cite{chen2024agentpoison} and MINJA~\cite{dong2026memory}. 
The experiments are conducted under the benign memory dilution setting described in Section~\ref{Memory_Dilution}.

The three attacks use different poisoning mechanisms: AgentPoison directly modifies the memory repository, MINJA introduces malicious records through query-based interactions, and \textsc{MemCollusion} introduces external content that must be observed, summarized, and persisted by OpenClaw. 
To isolate robustness to memory-level defenses from differences in poisoning difficulty, we allow each method to poison the agent's memory through its original mechanism. We then apply the same defense pipeline to the resulting memory records and compare the resulting relative reductions in ASR, thereby measuring how well each attack survives the defenses.

We evaluate two defenses: an LLM-based auditor that independently examines each stored memory and removes entries judged to contain unsafe or manipulative information, and A-MemGuard~\cite{wei2025memguard}, which detects anomalous reasoning through consensus validation and retains corrective lessons using a dual-memory structure. Results are averaged across backbone models and task domains.

As shown in Table~\ref{tab:memory_defense}, LLM Audit reduces the ASR of AgentPoison and MINJA by 89.2\% and 75.1\%, respectively, while A-MemGuard reduces them by 100.0\% and 89.9\%. In contrast, the ASR of \textsc{MemCollusion} decreases by only 3.7\% under LLM Audit and 14.9\% under A-MemGuard. 
These results demonstrate that distributing the target information across multiple individually plausible memories substantially improves the attack's ability to evade memory-level defenses.

\begin{table}[t]
\centering
\small
\resizebox{0.8\linewidth}{!}{
\begin{tabular}{llcc}
\toprule
\textbf{Group} & \textbf{Method} & \textbf{MSR} & \textbf{ASR} \\
\midrule

\multirow{2}{*}{Basic Baselines}
& Direct Query           & --   & 2.1  \\
& Random Split           & 36.1 & 29.2 \\
\midrule

\multirow{2}{*}{Component Ablations}
& Single Memory          & 91.7 & 6.3  \\
& Prompted Generator     & 34.4 & 37.5 \\
\midrule

\multirow{5}{*}{Single-Strategy Variants}
& Equivalence Bridging   & 69.8 & 64.6 \\
& Consensus              & 68.1 & 66.7 \\
& Authority Endorsement  & 74.7 & 70.8 \\
& Workflow Continuation  & 64.6 & 62.5 \\
& Validation Closure     & 72.9 & 64.6 \\
\midrule

Full Framework
& \textsc{MemCollusion}  & \textbf{79.5} & \textbf{77.1} \\

\bottomrule
\end{tabular}
}
\caption{Ablation results on DeepSeek-V4-Flash under the S4 saving setting, averaged across three task domains.}
\label{tab:ablation}
\end{table}

\subsection{Ablation Study}
We conduct an ablation study on DeepSeek-V4-Flash under the S4 saving setting, with results averaged across three task domains. The evaluated methods are divided into four groups. \textit{Basic Baselines} include \textit{Direct Query}, which directly presents the target behavior to the agent, and \textit{Random Split}, which splits the target information without enforcing our constraints or applying any construction strategy. \textit{Component Ablations} include \textit{Single Memory}, which selects one fragment from a six-fragment coalition and evaluates it in isolation, and \textit{Prompted Generator}, which directly prompts the base model without fine-tuning. \textit{Single-Strategy Variants} construct all candidate coalitions using only one of the five construction strategies. \textit{Full Framework} denotes \textsc{MemCollusion} with all components enabled. All variants use five attack attempts per case, matching the attack budget of \textsc{MemCollusion}.

As shown in Table~\ref{tab:ablation}, Direct Query achieves only 2.1\% ASR, while Random Split improves it to 29.2\%, indicating that fragmentation alone is insufficient. Single Memory obtains a high MSR of 91.7\% but only 6.3\% ASR, confirming that individual fragments remain ineffective in isolation. Prompted Generator reaches 37.5\% ASR, demonstrating the importance of generator fine-tuning. The five construction strategies individually achieve 62.5--70.8\% ASR, showing that each is effective but none matches the full framework. Overall, \textsc{MemCollusion} achieves the highest ASR of 77.1\%, validating the complementary benefits of structured constraints, diverse construction strategies, and fine-tuning.


\section{Limitations and Scope}
\textsc{MemCollusion} requires multiple memory fragments to be jointly available. Its effectiveness may decrease in traditional agents that retrieve only top-1 or small top-$k$ memories, such as EHRAgent~\cite{shi2024ehragent}. In contrast, modern agents typically have sufficient context capacity to directly incorporate multiple stored memories, making such compositional risks practically relevant. Our study focuses on this setting, represented by OpenClaw.

\section{Conclusion}
In this paper, we introduce \textsc{MemCollusion}, an automated framework for collusive memory poisoning attacks. It applies salami tactics to divide an adversarial objective into multiple memory fragments that appear benign individually but become harmful when used together. We define four constraints and five construction strategies, and fine-tune a generator to produce such memory coalitions automatically. Experiments on OpenClaw show that \textsc{MemCollusion} is effective across sessions and remains robust under benign memory dilution and memory-level defenses.

\bibliography{aaai2027}



\end{document}